\documentclass{article}

\usepackage{arxiv}

\usepackage[utf8]{inputenc} % allow utf-8 input
\usepackage[T1]{fontenc}    % use 8-bit T1 fonts
\usepackage{hyperref}       % hyperlinks
\usepackage{url}            % simple URL typesetting
\usepackage{booktabs}       % professional-quality tables
\usepackage{amsfonts}       % blackboard math symbols
\usepackage{nicefrac}       % compact symbols for 1/2, etc.
\usepackage{microtype}      % microtypography
\usepackage{lipsum}		% Can be removed after putting your text content
\usepackage{graphicx}
\usepackage{natbib}
\usepackage{doi}
\usepackage{color}

\title{Deep Reinforcement Learning for Optimal Stopping with Application in Financial Engineering}

\date{\today} 					% Or removing it

\author{ {Abderrahim Fathan, Erick Delage} \\
	Department of Decision Sciences\\
	HEC Montreal\\
	Montreal, Quebec, Canada, H3T 2A7 \\
	\texttt{abderrahim.fathan@gmail.com, erick.delage@hec.ca} \\
}

\renewcommand{\shorttitle}{Deep RL for Optimal Stopping}

\hypersetup{
pdftitle={Deep Reinforcement Learning for Optimal Stopping with Application in Financial Engineering},
pdfsubject={q-bio.NC, q-bio.QM},
pdfauthor={Abderrahim Fathan, Abderrahim Fathan},
pdfkeywords={optimal stopping, reinforcement learning, deep learning, financial engineering},
}

\usepackage{eufrak}
\usepackage{hyperref}
\usepackage{url}

\usepackage[utf8]{inputenc} % allow utf-8 input
\usepackage[T1]{fontenc}    % use 8-bit T1 fonts
\usepackage{hyperref}       % hyperlinks
\usepackage{url}            % simple URL typesetting
\usepackage{amsfonts}       % blackboard math symbols
\usepackage{nicefrac}       % compact symbols for 1/2, etc.
\usepackage{microtype}      % microtypography

\usepackage{algorithm}
\usepackage{algorithmic}
\usepackage{multirow}
\usepackage{lscape}

\usepackage[flushleft]{threeparttable}

\usepackage{subfig}
\usepackage{hyperref}
\def\Expect{{\mathbb E}}

\def\min{\mathop{\rm min}}
\def\max{\mathop{\rm max}}

\def\sup{\mathop{\rm sup}}

\newcommand{\V}[1]{{{\boldsymbol #1}}}

\newcommand{\quoteIt}[1]{``#1''}

\newif \ifFinal
\Finaltrue %this is for final version
\ifFinal
\newcommand{\EDmodified}[1]{{#1}}
\newcommand{\EDmodifiedB}[1]{{#1}}
\newcommand{\EDcomments}[1]{{}}

\newcommand{\removed}[1]{{}}
\else
\newcommand{\EDmodified}[1]{{#1}}
\newcommand{\EDmodifiedB}[1]{{\color{red} #1}}
\newcommand{\EDcomments}[1]{{\EDmodifiedB{Erick commented: #1}}}

\newcommand{\removed}[1]{{}}
\fi

\newcommand{\timeF}{{\mathfrak{t}}}

\usepackage{mathtools}
\begin{document}
\maketitle

\begin{abstract}
	%\lipsum[1]
	Optimal stopping is the problem of deciding the right time at which to take a particular action in a stochastic system, in order to maximize an expected reward. It has many applications in areas such as finance, healthcare, and statistics. In this paper, we employ deep Reinforcement Learning (RL) to learn optimal stopping policies in two financial engineering applications: namely option pricing, and optimal option exercise. We present for the first time a comprehensive empirical evaluation of the quality of optimal stopping policies identified by three state of the art deep RL algorithms: double deep Q-learning (DDQN), categorical distributional RL (C51), and Implicit Quantile Networks (IQN). In the case of option pricing, our findings indicate that in a theoretical Black-Schole environment, IQN successfully identifies nearly optimal prices. On the other hand, it is slightly outperformed by C51 when confronted to real stock data movements in a put option exercise problem that involves assets from the S\&P500 index. More importantly, the C51 algorithm is able to identify an optimal stopping policy that achieves 8\% more out-of-sample returns than the best of four natural benchmark policies. We conclude with a discussion of our findings which should pave the way for relevant future research.
\end{abstract}

% keywords can be removed

%%%%%%%%%%%%%%%%%%%%%%%%%%%%%%%%%%%
\section{Introduction}

%\EDcomments{In sections 1,2,3,4, I tried to be consistent in how we call the two actions. I think the right one action is \quoteIt{stop}, the other should be \quoteIt{continue}. So we should refer to stop time, instead of exercise time or execution time. }
%\EDcomments{My main comment for presentation is to use general optimal stopping problem description in sections 1, 2, 3, 4 sporadically inserting some details for option pricing as an illustrative example. I have edited the first paragraph as an example.}

We consider the problem of Optimal Stopping (OS) in a stochastic system, which can be described as follows: the system evolves from one state to another in discrete time steps up to some fixed horizon $T$. At each time step the decision maker has the option to stop the process or wait for a later step to do so. If he decides to stop, then he gets a reward that depends on the current state of the system. Otherwise, the decision maker does not receive any reward immediately, but can decide to stop at a future time step.  

In spite of its simplicity, the optimal stopping model is of use in many fields of application including asset selling, gambling, and sequential hypothesis testing. Recently, in \citep{ajdari2019towards}, it is used to determine when to stop the treatment of patients receiving fractioned radiotherapy treatments.   \citep{liyanage2019automating} proposes an OS framework to perform feature selection in the classification of urban issue requests on civic engagement platforms. \citep{dai2019bayesian}  combines Bayesian optimization and OS in the design of early-stopping strategies for the training of neural networks. %\cite{OSPCache} uses  OS to identify when the cached version of a webpage should be updated. 
Its most popular application is however in financial engineering. For example, a Bermudan option is a financial derivative product with a predetermined maturity deadline that will pay out at exercise time, chosen among a discrete set of time points, an amount that depends on the value of an underlying financial asset. Based on no-arbitrage theory, these options are usually priced according to the expected return achieved by an optimal exercise (i.e. stopping) strategy under an assumed martingale stochastic process, such as a Geometric Brownian Motion (GBM). Alternatively, the buyer of such an option will often seek to exercise it at the most profitable moment without exact knowledge of the stochastic dynamics of the underlying asset.

In stochastic control, one way to solve the optimal stopping problem is by using Approximate Dynamic Programming (ADP) \citep{bertsekas1996neuro} %,puterman2014markov} 
to approximate the value function that specifies the best expected reward one can receive starting from a given state. Once this is achieved, a greedy policy based on the approximate value function is expected to provide good decisions. In this paper, we apply some of the latest advancements in reinforcement learning (RL), which had great success in controlling several Atari 2600 games \citep{atari2013,mnih2015humanlevel,bellemare2017distributional,dabney2018implicit}, to address the optimal stopping problem. The original algorithms have been modified to adapt to time series data and a Long Short Term Memory (LSTM) \citep{hochreiter1997long} recurrent neural network is implemented to model long sequences and integrate history. Following the work by \citep{hessel2018rainbow}, we also combine three additional techniques in each of our three customized RL approaches, the first of which is Double Q-learning, first introduced in \citep{DoubleQL} to address the problem of over-estimation of action values, and has been subject to improvements in \citep{DRLwithDQL} to attain better performance. The second is the dueling architecture \citep{duelingDRL} that uses two separate estimators: one for the state value function and one for the state-dependent action advantage function, which leads to better policy evaluation in the presence of many similar-valued actions. The third is multi-step bootstrapping of targets \citep{de2018multi} which helps accelerate the propagation of newly observed rewards to earlier visited states and \EDmodified{balances} the bias-variance trade-off. %shifts
%\EDcomments{This paragraph seems outdated. We should be discussing DDQN, C51, IQN acronyms.}

In this work, we additionally perform the first comprehensive empirical study of the use of recent \EDmodified{deep} reinforcement learning algorithms to solve the problem of optimal stopping with application in option pricing and optimal exercising. We show, for the first time, that with \EDmodifiedB{well-designed} modifications to the original algorithms, \EDmodified{deep RL architectures such as
Double Deep Q-Network (DDQN) \citep{mnih2015humanlevel}, Categorical Distributional RL (C51) \citep{bellemare2017distributional} and Implicit Quantile Networks (IQN) \citep{dabney2018implicit} are able to identify policies that} achieve near optimal performance in terms of pricing and to outperform predictive financial models, such as the binomial tree model, in an option exercising problem with real US stock market data
. Furthermore, our experiments demonstrate that: (1) models based on deep reinforcement learning have a high ability to learn and adapt to stochastic environments with high volatility and randomness; 
%(2) RL is capable of learning near optimal policies on GBM models, and outperform strategies based on such model when employed with real assets, and (3) performance of Categorical Distributional RL (C51) \citep{bellemare2017distributional} and Implicit Quantile Networks (IQN) \citep{dabney2018implicit} degrades if n-steps value bootstrapping is included, and the dueling architecture is usually harmful to performance in those 2 contexts, \EDcomments{I don't get this (3). We gave up on evidence of this fact right? I strongly recommend to remove it.}
(2) C51 and IQN algorithms outperform \EDmodified{DDQN} in terms of performance, at a cost of more computation time; (3) C51 slightly outperforms IQN when confronted to real stock data movements, identifying an option exercise policy that achieves 8\% more out-of-sample returns than the best of four natural benchmark policies.

%especially when compute time is taken into consideration.

The rest of this paper is organized as follows: In Section \ref{sec:relWork}, we discuss related work to our problem, then define the problem in Section \ref{sec:model}. Section \ref{sec:arch} describes the three RL architectures that will be evaluated. Section \ref{sec:application} then presents two experiments involving financial engineering applications that compare the performance of RL to natural benchmarks. Finally, in Section \ref{sec:discussion} we conclude with a discussion of our findings which should pave the way for relevant future research.  Interested readers can find all our data, implementations, and experiments at  \href{https://github.com/osrlpaper/os\_rl\_papercode}{https://github.com/osrlpaper/os\_rl\_papercode}.

\section{Related work}\label{sec:relWork}

Among the state-of-the-art ADP approaches that solve the optimal stopping problem, one finds the  simulation-regression approach of (see \cite{Carrire1996ValuationOT,Longstaff01valuingamerican,935083}) which uses regression to approximate the optimal continuation value at each state of the system, and the martingale duality-based approach of \citep{desaiMartingale}. The latter relaxes the non-anticipativity requirement of the policy by allowing it to use future information, but on the other hand penalizes any policy that uses such information. Several other approaches have been derived from these two including \removed{\cite{roger:MCvaluation},} \cite{brown:InfoRelax}, and \cite{goldberg2018beating}. However, such numerical methods either suffer from the well-known curse of dimensionality, assume that the underlying stochastic model is known (or in the very least the states that makes it Markovian), or require the fine tuning of basis functions. 

On the RL side, 
\citep{becker2019deep} proposes an approach in which multilayer feed-forward neural networks are used. \citep{li2009learning} also applies the least-squares policy iteration RL method to the problem of learning exercise policies for American options and shows their good quality. \removed{\cite{even2003action} provides a model-based and a model-free variant of a RL action elimination method, based on learning an upper and a lower estimates of the value function. They further derive stopping conditions that guarantee that the learned policy is approximately optimal with high probability. In addition, \citep{buehler2019deep} proposes a RL framework for hedging a portfolio of derivatives, and \citep{lu2017agent} proposes a recurrent RL algorithm with evolutionary strategies for robo-trading of long and short positions. \citep{jiang2017deep} presents a RL framework for portfolio management using a deterministic policy gradient algorithm.} \citep{misic2018} on the other hand addresses the OS problem by constructing interpretable optimal stopping policies from data using binary trees. \cite{goel2017sample} proposes a model-free policy search method that reuses data for sample efficiency by leveraging problem structure to simultaneously learn and plan. On the other hand, \cite{yu2007q} introduces alternative algorithms to Q-learning for OS, which are based on projected value iteration, linear function approximation, and least squares. \citep{becker2019deep} propose a value-based reinforcement learning for OS learning from Monte Carlo samples, with application to derivative pricing. It is the closest paper to our work, and the reader is referred to it for more details of the problem setup.%\EDcomments{I am not sure I agree. The deep optimal stopping paper seems pretty close to our application to GBM.}
\citep{hu2019deep} consider the problem of ranking response surfaces as image segmentation, using feed-forward neural networks to approximate the value function. Reformulating the optimal stopping problem as a surface ranking problem, they apply this scheme to pricing Bermudan options.
\citep{chen2019zap} propose a Q-learning based algorithm for OS with an application to derivative pricing. In their paper, they prove convergence of the algorithm using ODE analysis, and also observe that it achieves optimal asymptotic variance.

To the best of our knowledge, our paper is the first to apply and compare the performance of Deep DQN \citep{atari2013, mnih2015humanlevel}, Categorical Distributional RL \citep{bellemare2017distributional}, and Implicit Quantile Networks \citep{dabney2018implicit} on optimal stopping problems.
%\EDcomments{I removed the old text since this paragraph was not improved on, see in tex file.}

%%%%%%%%%%%%
%[OPTIONAL] The  methods  described  here  could  be  adapted to generalize  to  more sophisticated agents that trade and optimize a portfolio, trade varying quantities of a security, allocate assets continuously or manage multiple asset portfolios. Besides American option prices, these methods can easily generalize to more exotic option prices, for which hedge parameters can be learned. These techniques can also be used to speed up calibration,  and to  audit  mark-to-market  structured  products. \EDcomments{Let's remove this optional remark.}
%%%%%%%%%%%%%

\section{Problem definition}\label{sec:model}

In this work we adopt the following notation:\\
\hspace{-0.5cm}\begin{minipage}[t]{0.5\textwidth}
\begin{itemize}
\item $\beta:$ the discount factor $\in$ \EDmodified{[0,1]}%; and $T:$ the horizon of the problem.
%\item $T:$ the horizon of the problem
\item $\mathcal{A}_t:$ the set of possible actions at time $t$%; and $\mathcal{S}\subseteq \mathbb{R}^L:$ the set of possible states.
%\item $\mathcal{S}\subseteq \mathbb{R}^L:$ the set of possible states
\item $\Omega \subseteq \mathcal{S}^{T+1}:$ the set of possible trajectories%; and $\Pi:$ the set of possible policies $\pi:\mathcal{S}\rightarrow \mathcal{A}$.
%\item $\Pi:$ the set of possible policies $\pi:\mathcal{S}\rightarrow \mathcal{A}$.
\end{itemize}
\end{minipage}
\begin{minipage}[t]{0.55\textwidth}
\begin{itemize}
\item $T:$ the horizon of the problem.
%\item $T:$ the horizon of the problem
\item $\mathcal{S}\subseteq \mathbb{R}^L:$ the set of possible states.
%\item $\mathcal{S}\subseteq \mathbb{R}^L:$ the set of possible states
\item $\Pi:$ the set of possible policies $\pi:\mathcal{S}\rightarrow \mathcal{A}$.
%\item $\Pi:$ the set of possible policies $\pi:\mathcal{S}\rightarrow \mathcal{A}$.
\end{itemize}
\end{minipage}
In particular, we let $\mathcal{A}_t:= \{\mbox{continue}, \mbox{stop}\}$ for $t<T$, and $\mathcal{A}_T:=\{\mbox{stop}\}$, and always assume that the time (and remaining time $T-t$) can be inferred from a state of $s$, i.e. $t=\timeF(s)$ iff $s$ is observed at time $t$.
The stopping time with policy $\pi$ is defined as: $ \tau_\pi =\min\{ t \in[0,\dots,T]\,:\, \pi(s_t) = \textnormal{stop}\}  \,.$
Our goal is to find the optimal policy, that maximizes the average discounted return received over all trajectories: $\pi^* = {\arg\max}_{\pi\in\Pi} \mathbb{E}\left[\beta^{\tau_\pi} g_{\tau_\pi}(\V{s}_{\tau_\pi:T})\right]$,
where $\V{s}_{t:T}$ is shorthand for the sub-trajectory $[s_t,\dots,s_T]$ and the expectation is taken based on the distribution of $\V{s}_{0:T}$. Furthermore, $g_t(\V{s}_{t:T})$ refers to the payout received when stopping at time $t$ under any trajectory with a tail trajectory matching $\V{s}_{t:T}$. 
Alternatively, one can also define 
\[\pi^*(s_t) = \left\{ \begin{array}{cl}\mbox{stop}&\mbox{if $\Expect[\beta^t g_t(s_{t:T})|s_t]\geq \Expect[\beta^{\tau_\pi^{t+1}} g_{\tau_\pi^{t+1}}(\V{s}_{\tau_\pi^{t+1}:T})|s_t]$}\\ \mbox{continue} & \mbox{otherwise}\end{array} \right.\]
%{\arg\max}_{a\in\mathcal{A}_t} \mathbb{E}\left[\beta^{\tau_\pi^t} g(\tau_\pi^t,\omega)|X_t\right]\,,\]
where $\tau_\pi^t = \min\{ t' \in[t,\dots,T]\,:\,\pi(s_{t'}) = \textnormal{stop} \}$.

% During training, this translates to:
%$$ \max_{\pi\in\Pi} \frac{1}{|\Omega|} \sum_{\omega} \beta^{\tau_\pi-1} g(\tau_\pi,\omega) $$

In the case of a Bermudan option pricing or exercising problem, $s$ includes information about the current value of the financial asset, which can be recovered through some $S(s)$, and $g_t(\V{s}_{t:T}):=\max(0,\,S(s_t)-K)$ with strike price $K$ for a call option or $g_t(\V{s}_{t:T}):=\max(0,\,K-S(s_t))$ for a put option. In other words, a call option will pay the difference between the value of the asset and the strike price if it is positive, and the opposite occurs for the put option. When $K$ is set to be $S(s_0)$, the option is said to be at-the-money. 

We also wish to emphasize that our payout function is flexible enough to model any general payout $h(t,\V{s}_{0:T})$ as long as the information about $\V{s}_{0:t-1}$ (or some \quoteIt{sufficient statistics}) is included in $s_t$, obtaining $g_t(\V{s}_{t:T}):=h(t,[\V{s}_{0:t-1}(s_t),\; \V{s}_{t:T}])$. For example, in an exercise problem involving a call option, one might be instead interested in maximizing $\Expect[\max(0,\,S(s_{\tau_\pi})-K)/\sup_t \max(0,\,S(s_t)-K)]$. This can easily be implemented by including in the state $\theta_t = :\max(\theta_{t-1},\max(0,\,S(s_t)-K))$, and defining $g_t(\V{s}_{t:T}):=\max(0,\,S(s_{t})-K)/\max(\theta_t, \sup_{t'>t} \max(0,\,S(s_{t'})-K))$ with $\beta=1$.

Finally, one can show that the sequential decision making problem described above can be reformulated in standard reinforcement learning notation as:
\begin{eqnarray}
Q(s,a):= \left\{\begin{array}{cl}r(s,\mbox{stop}) & \mbox{if $a=\mbox{stop}$}\\ r(s,\mbox{continue})+\beta\Expect[\max(Q(s',\mbox{stop}),\,Q(s',\mbox{continue}))] & \mbox{ otherwise.}\end{array}\right. \label{eq:Qfunction}
\end{eqnarray}
where $r(s,\mbox{continue}):=0$ while $r(s,\mbox{stop}):=g_{\timeF(s)}(\tilde{\V{s}}_{\timeF(s):T})$ for a random trajectory $\tilde{\V{s}}_{0:T}$  with $\tilde{\V{s}}_{\timeF(s)}=s$. In what follows, we provide for the first time customized implementations of a number of state-of-the-art deep RL algorithms to this most general form of the OS problem. In particular, the original Double Q-learning \citep{mnih2015humanlevel} only uses plain fully-connected layers that exhibit low performance when applied on strategic games with long time dependencies. In our implementation, we integrated LSTM in order to learn the representation of states, aggregate partial information from the past, and capture long-term dependencies in our sequential data.

%\EDcomments{I don't like this sentence: It is also worth mentioning that the optimal stopping problem is different from regression and classification in that the former is a more complex problem than just approximating some functions or labeling data.}
It is also worth mentioning that OS problems cannot be exactly cast as a regression of the optimal stopping time, or the classification of $X$ as either \quoteIt{stop} or \quoteIt{continue} given that the trajectories are unlabeled, and that the consequences of continuing are delayed.

%%%%%%%%%%%%%%%%%%%%%%%%%%%%

\section{Architectures}\label{sec:arch}

We implemented three deep reinforcement learning algorithms to identify the optimal policy and value of the OS problem: double deep Q-Learning (DDQN), categorical distributional RL (C51), and implicit quantile networks (IQN). In short, DDQN attempts to learn an optimal action-value Q-network (as defined in \eqref{eq:Qfunction}), while C51 and IQN aim at learning the full distribution of the total discounted reward, i.e. the optimal stopping value $\beta^{\tau_{\pi^*}} g_{\tau_{\pi^*}}(\V{s}_{\tau_{\pi^*}:T})$, given $s$. For conciseness, we push the pseudo-code description of DDQN and C51 to Appendix \ref{app:alg}.  \removed{Algorithm \ref{alg:doubleQLearning} shows our implementation of double deep Q-Learning. C51 and IQN versions were similarly developed by adapting the action-value Q-network to learn the value distribution of returns, and to additionally approximate its quantile function, respectively, instead of only expected values.} 
We also refer interested readers to \cite{mnih2015humanlevel},  \citep{bellemare2017distributional}, and \citep{dabney2018implicit} for additional details on the original implementations of these three approaches. Since we are dealing with time series of varying lengths (they end when the agent stops the process), the core of our model uses a dynamic LSTM layer\EDmodified{: specifically, we used three layers with cells of size 512.}
Our architecture integrates two neural networks: a primary network to choose an action given the current state, and a target network that generates the target Q-value for that action. Adam gradient descent \citep{kingma2014adam} was used to optimize our networks using Huber loss as our temporal difference error.

\EDmodified{During each episode of training}, we first decide whether the episode will employ a random policy or not. If it is random, then a random stopping time is chosen. Otherwise, the policy learned so far is used throughout the episode. The probability of employing a random policy $\epsilon$ is annealed from 1 to 0.01 over time, and no random action is taken during \EDmodified{validation or} test. We observed that this approach improved learning efficiency when compared to $\epsilon$-greedy policies since it avoided unnecessary use of the neural network in trajectories where a random action eventually ends up being taken. Our exploration strategy also quickly provided to the agent a more diversified set of experiences to learn from, with a stopping time that is uniformly distributed over the horizon compared to $\epsilon$-greedy which had a bias towards stopping early.  Given the nature of our application, it would be in theory possible to dismiss exploration altogether if the whole trajectory (even passed the stopping time) was used in training. However, we found that learning only from the part of the trajectory that precedes stopping time together with some annealed exploration improved the quality of final policies. We suspect that this is due to the fact that the neural network's predictive power ends up focusing more on relevant states. We also consider that the trajectory past the stopping time might be unobservable in some applications, e.g. the secretary problem.
%to future data is possible, another possible approach to follow would be to keep learning from trajectories as monte carlo simulations without exploration, or by exploiting future information while penalizing the RL agent for using it. 
%Our proposed approach remains, however, more flexible, more generalizable, and simply follows the current practice in RL.
%[EXPERIMENTS STILL RUNNING, WE'LL SEE IF THIS WORKS. THOUGH THE NEED OF RL agents FOR BAD/GOOD decisions (errors) during training is still not very clear to me]
 
It is worth mentioning that, unlike in the original \EDmodified{DDQN} algorithm \citep{mnih2015humanlevel}, we take into consideration the sequential nature of data.
%\EDcomments{I don't understand the next sentence.}
Hence, when sampling \EDmodified{a mini-batch of buffers B, we provide a sequence of T time steps}
%, we provide the sequence B of T time steps,
and a maximum of $T*$ batch-size sequence transitions are used in every mini-batch, depending on the stopping time of every episode ($(T-n+1)*$ batch-size in case of n-step bootstrapping).\EDcomments{Is the previous paragraph clear ?}
%
%\EDcomments{Confusion about fact that DDQN improved with dueling architecture. Explain that no statistically significant improvement were observed from 7-step.}
Furthermore, the maximum dueling architecture and multi-step bootstrapping of 7-steps \citep{de2018multi} have been integrated and optimised into our version of DDQN, however, they degraded the results for IQN and C51 during tests and hence their use was omitted in all versions. 
%Table \ref{tab:hyperparams-table} shows the best hyperparameters optimised seperately for DDQN, C51, and IQN. %For simplicity and clarity Algorithm. \ref{alg:doubleQLearning} only shows a 1-step look-ahead version. 3 values for   bootstrapping were tested (3, 5 and 7) and we found 7 to perform better.

%Table \ref{tab:hyperparams-table} shows the best hyperparameters optimised seperately for DDQN, C51, and IQN versions.%\EDcomments{On what type of data ? Abderrahim: I need to update it. I will have to seperate hyperparams for both data sets}

Finally, we added a dropout wrapper around the LSTM and fully connected layers, with a drop probability of 20\%, to reduce the risk of over-fitting during training. Soft-updates ($\tau=0.001$) of the target network were implemented and tested, where the network is smoothly and gradually updated in contrast to hard updates that assign the whole online network to the primary network at each update. Overall, while hard update appeared more stable and better performing when trained with synthetic data in Section \ref{sec:BSpricing}, soft-updates were the favoured configuration for real data training (in Section \ref{sec:sp500}) for both C51 and IQN. 

All the code was implemented in tensorflow 1.14 using, among others,  CudnnCompatibleLSTMCell on a GPU, which is 3-5x faster than normal LSTM  implementations, and is platform-independent. To further accelerate learning, We first anneal $\epsilon$ rapidly to allow the algorithm to learn from more meaningful samples, then we decelerate the annealing speed through time. This has proven to be more efficient during our experiments.

%\begin{figure}
%\begin{minipage}[t]{0.6\linewidth}
%  \centering
%\includegraphics[height=5cm]{./figures/Probability_of_Random_Action_over_time.png}
%\caption{Frequency of random episodes over time.}
% \label{frequency_of_random_episodes_over_time}
%\end{figure}

%To accelerate learning in our case, We first anneal $\epsilon$ rapidly to allow the algorithm to learn from more meaningful samples, then we decelerate the annealing speed through time. This has proven to be more efficient during our tests.

% \usepackage{booktabs}
% \usepackage{multirow}
\removed{
\begin{table}[]
\centering
\caption{Hyperparameters of the different RL versions.}
\label{tab:hyperparams-table}
\begin{tabular}{@{}ccl@{}}
\toprule
Task                     & Algorithm & \multicolumn{1}{c}{Hyperparameters}                                          \\ \midrule
\multirow{3}{*}{GBM}     & DDQN      & learning-rate=0.0001  + batch-size=128 + $C$=3000  + $U$=300  \\ \cmidrule(l){2-3} 
                         & C51       & learning-rate=0.0025  + batch-size=64  + $C$=3000  +  $U$=30  \\ \cmidrule(l){2-3} 
                         & IQN       & learning-rate=0.00005 + batch-size=128 + $C$=3000  + $U$=1000 \\ \bottomrule
\multirow{3}{*}{S\&P500} & DDQN      & learning-rate=0.005  +  batch-size=64  + $C$=10000  + $U$=300 \\ \cmidrule(l){2-3} 
                         & C51       & learning-rate=0.0025  +  batch-size=64  + $C$=3000  +  $U$=30 \\ \cmidrule(l){2-3} 
                         & IQN       & learning-rate=0.0025  + batch-size=64  + $C$=3000  + $U$=100  \\ \midrule
\end{tabular}
\end{table}
}

%%%%%%%%%%%%%%%%%%%%%%%%%%%%%

\section{Empirical Results}\label{sec:application}
%\EDcomments{I would present section on GBM before real stock data}
%\EDcomments{The next paragraph is implementation details and should be postponed to the numerical experiments section.}
 %$g(\tau_\pi,\omega)$ is always positive as decision to exercise option is only made in case of positive payoff.

\newcommand{\Train}{{Training} }
\newcommand{\ValidHP}{{Valid\_HP} }
\newcommand{\ValidM}{{Valid\_Model} }

In this section, we assess the performance of three different RL algorithms on two financial engineering problems. In the first one, RL is used to price a Bermudan put option in a context where the underlying stock dynamics are assumed to be known. This is a case where a unique price exists and can be computed numerically by employing approximation methods such as binomial tree models. We are therefore able to compare the performance of RL to a ground truth which will validate the potential of C51 and IQN at identifying truly optimal policies. The second setting involves an optimal exercise problem where the underlying stock's dynamics are unknown and based purely on historical data. We will show that in this real world setting, state of the art methods like C51 can learn policies that significantly outperform traditional benchmarks out-of-sample.

In both applications, the state $s\in\mathcal{S}$ will be  defined as a sequence of $L=15$ scalar values (history of prices), concatenated with the amount of remaining time ($T-t$) to maturity of the option and the relative position of the stock value compared to strike price $\beta^{t}\max(K - S(s_t),0)-\max(S(s_t)-K,0)$), a feature that either returns the discounted reward that will be received (if strictly positive), or otherwise returns how far the stock is from the strike price.  This makes the real size of states fed to neural networks $L+2$. In order to warm start the LSTM, each episode is started 12 days earlier while the policy is only implemented from day 1.
%, before, with $L=15$ to account for the 15 most recent days and 12 days further spread historical days. Here, $L$ is 15 days, and 12 history states (or a total of 26 previous days) were used to warm-up the LSTM state before inference starts at the beginning of each episode.
Finally, we limited the number of epochs of training to 5 to avoid overfitting and also to limit computation time. In the special case of C51, in order to have comparable training times, C51 was trained on a subset of only 48 trajectories (instead of 160) in Section \ref{sec:BSpricing}, while in Section \ref{sec:sp500} it was trained for only three epochs. 

Our experiments will systematically involve three steps of execution. First, we calibrate the hyper-parameters of each algorithm using a training (\Train) and validation set (\ValidHP). Once the optimal setting is found, we employ a second validation set (\ValidM) in order to assess in an unbiased way which algorithm is best performing and finally test this best performing algorithm on a reserved set of test data. This process allows us to make claims about the statistical significance of our results. Performance of the RL policies will be compared to three natural benchmarks: \quoteIt{Rand} which chooses uniformly at random the exercise among the $T$ alternatives, \quoteIt{First} and \quoteIt{Last} which exercise respectively on the first $\tau=0$ and last $\tau=T$ days, and a binomial tree model (B.M.) that is either calibrated on the true stock dynamics (in Section \ref{sec:BSpricing}) or on the available set of $L$ historical prices (in Section \ref{sec:sp500}).

\subsection{Bermudan option pricing under Black-Schole setting}\label{sec:BSpricing}
%\EDcomments{This is a section about pricing. Here the stochastic model is always given, it needs to be a martingale. We don't care about the policy but really the price. You should focus on these measures. Abderrahim: I tried to adapt by talking about pricing}

Our first experiment consists of a classical Black-Schole option pricing problem. When a financial asset is assumed to behave according to a Geometric Brownian Motion (GBM) model, it is well known that, in order to avoid giving rise to arbitrage opportunities, a financial derivative of this asset needs to be priced according to the optimal expected revenue that can be obtained under the GBM's so called risk neutral martingale measure. Here, we focus on the case of pricing an at-the-money Bermudan put option with daily exercise opportunities (often used as a proxy for pricing American options), where the daily discretized risk neutral measure takes the form of $S_t = S_{t-1} e^{(r - \frac{\sigma^2}{2}) \Delta t + \sigma \sqrt{\Delta t} \varepsilon}$, with $r$ as the risk-free continuously compounded yearly interest rate, $\sigma$ as the volatility of the asset, $\varepsilon$ as the standard normal distribution, and $\Delta t$ as the amount of time elapsed between $t-1$ and $t$. Specifically, in our experiments, we let $S_0=1$, $\sigma = 20\% $, $\Delta t=1/252$, $r = 5\%$, and the horizon $T=38$. Similarly as in \cite{li2009learning}, RL will consider a discount factor of $\beta=e^{-r\Delta t}$ which effectively prices an option that pays $e^{-rt\Delta t}\max(0,S_0-S_t)$ at exercise time.
%\EDcomments{Previously as: \quoteIt{Specifically, in our experiments, we let $\sigma = 20\% $, $\Delta t=1/252$, and $r = 5\%$, which implies that $\beta=e^{-0.05/252}$.} But I think this does not make sense}
Hence, in this experiment, we train the three RL algorithms on simulated trajectories in order to use the expected reward from the best trained model as an estimation of the arbitrage-free option price (AFOP). While such a price can be obtained with high precision much more efficiently using binomial tree models (B.M.), our aim is to verify whether modern RL algorithms are mature and flexible enough to reach optimality and retrieve such a price.

%Another test was similarly performed on data generated according to the Geometric Brownian Motions model (GBM). Here, we suppose the stock price $S_t$ to follow a GBM at every discrete time step $t$, thus satisfies the following equation: $S_t = S_{t-1} e^{(r - \frac{\sigma^2}{2}) \Delta t + \sigma \sqrt{\Delta t} \varepsilon}$, where $r$ is the risk-free continuously compounded yearly interest rate, $\sigma$ captures the volatility of the stock , is the standard normal distribution, and $\Delta t$ is the amount of time elapsed between $t-1$ and $t$. Specifically, in our experiments, we let $\sigma = 20\% $, $\Delta t=1/252$, and $r = 5\%$, which implies that $\beta=e^{-0.05/252}$.\EDcomments{I don't get why you are using a discount factor. I am pretty sure that here we need $\beta=1$. If you did employ $\beta=e^{-0.05/252}$, a solution here would be to simply let $r=0\%$ and $\beta=1$}

%In this experiment, the training set is composed of 200 stock trajectories of 928 days each (a total of 135600 episodes for training, 24000 for validation), and an additional 200 paths (159600 different episodes) for test, generated with a different seed. Table \ref{tab:gbm-table} shows the different results.

In this experiment, the \Train set is composed of 160 sampled trajectories of 928 days each, from which are drawn 135600 episodes used in training. The \ValidHP set (for hyper-parametrization) consists of 40 independently and identically drawn trajectories of 928 days (24000 episodes), while the \ValidM (for algorithm selection) and Test set consist of 200 and 400 i.i.d. trajectories over 928 days respectively (159600 and 319200 episodes). 
We refer the reader to Appendix \ref{app:hyper}, which descibes the final choice of hyper-parameter values.
\removed{Table \ref{tab:hyperparams-table} summarizes the hyper-parameters.
%\EDcomments{Abderrahim, could you replace the XXX with actual numbers.}  
}

Table \ref{tab:gbm-table} presents the results for this experiment. Looking at the numbers, we observe that both C51 and IQN achieve high Expected Reward (ER) in training and both steps of validation. While C51 appeared to be the best performing approach on \ValidHP, we suspect that the selection of hyperparameters overfitted the \ValidHP set given that 1) it outperformed the theoretically optimal policy generated by the binomial tree model; 2) the performance degraded when validating on \ValidM set. Given its better performance in \ValidM, IQN was selected for the final out-of-sample test where it estimates a AFOP of $0.0284\pm 0.0001$ compared to a ground truth of $0.0283\pm0.0001$. This confirms that the resulting IQN exercise policy is statistically equivalent to the theoretical policy.

Overall, we can conclude that, despite the context of high stochasticity of GBMs, RL models such as IQN are flexible enough to learn optimal exercise policies. This shows the high potential of RL algorithms to replace conventional approaches in situations where the dynamics of the risk neutral martingale require a large state space in order to become Markovian, and should be easier to adapt to situations were the market is incomplete or stock dynamics are unknown. On the other hand, one needs to be aware of the heavy computational burden imposed by current state-of-the-art RL algorithms. Beyond requiring substantial training time due to their model-free nature, the selection of best performing hyper-parameters is still more of an art than a science. In particular, we observed that regions of best performing hyperparameter values were sensitive to factors such as the number of trajectories and epochs that were used.

%Here, we only report 1-step predictions since in the context of GBMs, predicting several days in advance does not make sense and leads our algorithms to underperfom or get stuck in local minima. Also, since reaching a minimum or a maximum at the end of a $L$ window frame is difficult, Min and Max would generally finish up having the same performance as Last. 

% \usepackage{booktabs}
% \usepackage{multirow}
\begin{table}[]
\centering
\caption{Performance of the 3 versions of RL vs Baselines for GBM data.}
\label{tab:gbm-table}
\begin{tabular}{@{}ccccccccc@{}}
\toprule
\multicolumn{2}{c}{Data $\backslash$ Method}           & DDQN                     & C51                      & IQN             & Rand                     & Last                     & First                    & B.M.                     \\ \midrule
                               & ER         & 0.0263                   & 0.0267                   & \textbf{0.0268} & 0.0229 & 0.0263 & 0.0160 & 0.0267                   \\ \cmidrule(l){2-9} 
\multirow{-2}{*}{\Train}     & Time (sec) & \textbf{0.4666}          & 1.3029                   & 0.6668          &  &  &  &  \\ \midrule
Valid\_HP                      & ER         & 0.0279                   & \textbf{0.0280}          & 0.0275          & 0.0234                   & 0.0279                   & 0.0167                   & 0.0276                   \\ \midrule
                               & ER         & 0.0270                   & 0.0273                   & \textbf{0.0275} & 0.0236                   & 0.0271                   & 0.0163                   & 0.0275                   \\ \cmidrule(l){2-9} 
\multirow{-2}{*}{Valid\_Model} & CI         & 0.0002                   & 0.0002                   & 0.0001          & 0.0001                   & 0.0002                   & 0.0001                   & 0.0001                   \\ \midrule
                               & ER/AFOP    & 0.0282 & 0.0283 & \underline{\textbf{0.0284}} & 0.0243                   & 0.0282                   & 0.0168                   & 0.0283                   \\ \cmidrule(l){2-9} 
\multirow{-2}{*}{Test}         & CI         & 0.0001 & 0.0001 & \underline{0.0001}          & 0.0001                   & 0.0001                   & 0.0001                   & 0.0001                   \\ \bottomrule
\end{tabular}

\begin{tablenotes}
      \small
      \item Best values are marked in \textbf{bold}. Training time is per episode (on a Titan X GPU). CI refers to a 90\% confidence interval. Out-of-sample performance of RL model selected with \ValidM is underlined.
\end{tablenotes}

\end{table}

%% S&P 500 experiments

\subsection{S\&P 500 stock data}\label{sec:sp500}

In this section, we consider the optimal exercise problem of a Bermudan option. In particular, we consider the distribution of $T$ days stock trajectories in which one first draws the stock randomly from 111 stocks that compose the S\&P 500 index, and a random date on the period 2014-03-27 to 2019-12-10. The \Train set considers trajectories from a subset of 60 different stocks and dates from the period 2014-03-27 to 2016-03-29 (733 trading days), \ValidHP considers the same set of stocks with period 2016-03-29 to 2017-11-10 (183 days). The \ValidM set considers 51 other stocks over the period 2014-03-27 to 2017-11-10. Finally, the test set is composed of all 111 stocks over a \quoteIt{future} period 2017-11-11 to 2019-12-10 (522 working days). In order for the policies to treat similarly stocks with different starting price, we focus on the task of maximizing the Expected Relative Option Payout (EROP) for an at-the-money put option with horizon $T=38$: i.e. $g_t(\V{s}_{t:T}):=(1/S(s_0))\max(0,S(s_0)-S(X_t))$. Once again, the performance is compared to Rand, First, and Last policies, while B.M. captures the optimal policy for a GBM calibrated on the last $L$ days. The same discount rate of $\beta=e^{-0.05/252}$ was used. Finally, we let the reader refer to Table \ref{tab:hyperparams-table} to find the best hyperparameters found using \ValidHP for each algorithm.

Table \ref{tab:sp500-table} shows the performance of the 3 RL algorithms against the four benchmarks. We can see that both C51 and IQN outperform DDQN in the \ValidHP set, this is confirmed in the \ValidM set which points to C51 as the best model to recommend for out-of-sample tests,  although IQN holds a tight second place. The Test set demonstrates that the best IQN outperforms significantly the four benchmarks in terms of Expected Reward (and EROP). Indeed, it achieves on average a 2.91\% relative option payout compared to exercising on the last day which achieves 2.17\%, and the binomial tree model approach that achieves 2.53\%. The table also presents Expected Option Return (EOR) which accounts approximately for the return on investment when implementing each policy assuming that the option is priced based on a GBM risk neutral measure calibrated on the recent history. Specifically, we see that C51 achieves a 22.0\% return on average which is 8\% higher than any of the competing classical benchmark.

%outperforms all the other algorithms, including C51 and IQN which both fare better than the remaining baselines. DDQN performs 6.3\% better than IQN; the second best algorithm on the valid set, and about 27.1\% better than B.M. On the test set, DDQN performs 7.3\% better than IQN, and around 12\% better than Min; the best performing among the baselines. Figure \ref{Q_and_loss_over_time} shows the improvement of performance over epochs for both the loss and the Q functions. From a financial point of view, we also include 2 additional performance metrics: The percentage of option return ((option payoff - option price)/option price), and the percentage of payoff return or stock-price normalized option payoff (option payoff/stock price). Option prices were calculated using our B.M. algorithm. The latter metrics further confirm the good performance of our approach. 

We wish to emphasize that, throughout our extensive set of experiments, including unreported experiments with stocks which dynamics followed a more sophisticated Generalized AutoRegressive Conditionally Heteroscedastic (GARCH) stock model, we observed that IQN has the ability to rapidly fit the training data, although this can in some cases lead to overfitting. Also, during our experiments, we noted that DDQN was 1.2-1.5x faster than IQN and around 2-4x faster than C51 depending on machine configuration, the type of GPU and available memory. Finally, IQN consumes considerably more memory than DDQN. These are all characteristics that are worth taking into account when choosing the right RL approach.

\begin{table}[]
\centering
\caption{Performance of the 3 versions of deep RL vs Baselines for S\&P 500 data}
\label{tab:sp500-table}
\begin{tabular}{@{}ccccccccc@{}}
\toprule
\multicolumn{2}{c}{Data $\backslash$ Method}                                                                                          & DDQN            & C51              & IQN             & Rand                     & Last                     & First                    & B.M.   \\ \midrule
Training (60 stocks)                                                                                                & ER/EROP   & 0.0290          & 0.0326           & \textbf{0.0336} & 0.0249 & 0.0281 & 0.0174 & 0.0270 \\ \midrule
Valid\_HP (60 stocks)                                                                                               & ER/EROP   & \textbf{0.0156} & 0.0126           & 0.0147 & 0.0130                   & 0.0118                   & 0.0082                   & 0.0123 \\ \midrule
                                                                                                                    & ER/EROP   & 0.0274          & \textbf{0.0289}  & 0.0285 & 0.0245                   & 0.0272                   & 0.0166                   & 0.0256 \\ \cmidrule(l){2-9} 
\multirow{-2}{*}{\begin{tabular}[c]{@{}c@{}}Valid\_Model\\ (51 other stocks)\end{tabular}}                          & CI   & 0.0003          & 0.0004           & 0.0003          & 0.0003                   & 0.0004                   & 0.0003                   & 0.0004 \\ \midrule
                                                                                                                    & ER/EROP   & 0.0285          & \underline{\textbf{0.0291}}  & 0.0281          & 0.0265                   & 0.0271                   & 0.0175                   & 0.0258 \\ \cmidrule(l){2-9} 
                                                                                                                    & CI   & 0.0003          & \underline{0.0004}           & 0.0003          & 0.0004                   & 0.0004                   & 0.0003                   & 0.0004 \\ \cmidrule(l){2-9} 
                                                                                                                    %& EROP & 2.85\%          & \underline{\textbf{2.91\%}}  & 2.81\%          & 2.65\%                   & 2.71\%                   & 1.75\%                   & 2.58\% \\ \cmidrule(l){2-9} 
                                                                                                                    %& CI   & 0.03\%          & \underline{0.04\%}           & 0.03\%          & 0.04\%                   & 0.04\%                   & 0.03\%                   & 0.04\% \\ \cmidrule(l){2-9} 
                                                                                                                    & EOR  & 17.1\%         & \underline{\textbf{22.0\%}} & 17.6\%         & 8.8\%                   & 13.9\%                  & -32.6\%                 & 5.3\% \\ \cmidrule(l){2-9} 
\multirow{-6}{*}{\begin{tabular}[c]{@{}c@{}}Test \\ (all stocks combined \\ on a future period)\end{tabular}} & CI   & 1.2\%          & \underline{1.6\%}           & 1.3\%          & 1.4\%                   & 1.6\%                   & 1.2\%                   & 1.5\% \\ \bottomrule
\end{tabular}

\begin{tablenotes}
      \small
      \item Best values are marked in \textbf{bold}. CI refers to a 90\% confidence interval. Out-of-sample performance of RL model selected with \ValidM is underlined.
\end{tablenotes}

\end{table}

%%%%%%%%%%%%%%%%%%%%%%

%%%%%%%%%%%%%%%%%%%%%%%%%%%%%%%

% \usepackage{booktabs}
% \usepackage{multirow}

%%%%%

%\section{Contributions}

\section{Concluding Discussion}\label{sec:discussion}

Solving the problem of optimal stopping in finance where data is known to have a high degree of randomness (unpredictable) is both a notoriously challenging and intriguing task. 
In this paper, we demonstrated the ability of three variants of deep reinforcement learning algorithms (DDQN, C51, and IQN) to learn simply from real historical stock price observations complex stopping time policies in the presence of uncertainty, volatility, and \EDmodifiedB{non-stationarities}. Despite being more difficult to employ and requiring a more significant computional investment than traditional off-the-shelf methods, our experiments present empirical evidence that these deep RL algorithms are flexible enough to retrieve optimal policies in context where these can be computed exactly (option pricing under GBM dynamics), and to significantly out-perform off-the-shelf methods when the dynamics of the underlying stochastic system are both unknown and likely to violate simplifying Markovian assumptions. In particular, distributional IQN and C51 are able to learn the value distribution of option returns and rise up as the favoured algorithms to employ in practice, with a strong preference for C51 when computation time is less of an issue.

\removed{Further improvements to these three algorithms could certainly be investigated. In particular, in the option pricing application, one could boost performance by accessing additional side-information about the financial markets (observations of market indexes or stock market, financial and business news) or about the underlying asset (e.g. financial statements). There are also several opportunities for improving these networks. A WaveNet model \citep{oord2016wavenet} could be used to replace the slower LSTM.  Convolutional layers with filters could also be added to the network. These filters can then be interpreted as feature maps that hold technical significance (as is often the case in image processing applications), which could help with the interpretability of the RL policy. Finally, one could also integrate some \emph{learning to search} techniques to improve training time and stability.}

In closing, it is worth mentioning that our experience of hyper-parameters tuning taught us that it is demanding and fragile, often requiring us to re-align the search grid the moment that problems are slightly modified.
We also observed in our experiments with real stock data, that it could be beneficial to avoid shuffling the episodes during training with the effect of improving the out-of-sample performance in periods that are chronologically close to the last episodes that were trained on. This idea could potentially be useful in online learning, when the underlying process is non-stationary, since it implicitly fine-tunes the algorithm according to the most recent data. We believe these constitute two important directions of future investigation.
%%%%%%%%%%%%%%%%%%%%%%%%%%%%%%%%%%%
\bibliographystyle{unsrtnat}
\bibliography{references}  %%% Uncomment this line and comment out the ``thebibliography'' section below to use the external .bib file (using bibtex) .





% \end{thebibliography}

%%%%%%%%%%%%%%%%%%%%%%%%%%%%%%%%%%%
\newpage
\appendix

\section{Pseudo-code for DDQN and C51 algorithms}\label{app:alg}
\subsection{Customized Double Deep Q-Learning Algorithm}\label{app:ddqn}

\begin{algorithm}[h]%[tb]
   \caption{Customized Double Deep Q-Learning Algorithm, a.k.a. DDQN}
   \label{alg:doubleQLearning}
\begin{algorithmic}[1]
   \STATE {\bfseries Inputs:} A set of $M$ episodes $\{\V{s}_{0:T}^i\}_{i=1}^M$
   \STATE {\bfseries Initialize:} replay memory $D$ to capacity $C$ of episodes
   \STATE {\bfseries Initialize:} action-value function $Q$ with random weights $\theta$
   \STATE {\bfseries Initialize:} target action-value function $\widehat{Q}$ with weights $\phi=\theta$
   \FOR{episode $i =1$ {\bfseries to} $M$}
   \STATE {\bfseries Initialize:} episode buffer $B$ to capacity $T$ \label{alg:searchA}
   \STATE With probability $\epsilon$ episode is random and a uniformly random day to stop $t_{stop}$ is selected 
   \FOR{$t=0$ {\bfseries to} $T$}
   %\STATE With probability $\epsilon$ select a random action $a_t$
   %\STATE otherwise select $a_t=\arg\max_a{Q(s_t,a;\theta)}$
   %%%
   \IF{episode is random}
   %\STATE Set $y_j=r_j$
   %%%%%%%%
   \STATE Set $a_t=\left\{\begin{array}{cl}\mbox{continue}&\mbox{ if $t<t_{stop}$}\\\mbox{stop}&\mbox{ otherwise.}\end{array}\right.$
   %%%%%%%%
   \ELSE
   \STATE Set $a_t=\arg\max_a{Q(s_t^i,a;\theta)}$ \label{alg:searchB}
   \ENDIF
   %%%
   %\STATE select $a_t=\arg\max_a{Q(s_t,a;\theta)}$
   \STATE Execute action $a_t$ and observe reward $r_t$, store transition $(s_t^i, a_t, r_t, s_{t+1}^i)$ in $B$
   \IF{($a_t=\mbox{stop}$ or $t=t_{stop}$)}% and $K-x_{k+t-1}>0$}
   \STATE Exit \textbf{for} loop
   \ENDIF
   \ENDFOR
   \STATE Store episode buffer $B$ in replay memory $D$. If $D$ full, drop the oldest episode \label{alg:searchC}
   \STATE Sample a random mini-batch of buffers $B$ of sequence transitions $(s_j, a_j, r_j, s_{j+1})$ from $D$
   \STATE Set $y_j=\left\{\begin{array}{cl}r_j&\mbox{ if $a_j=\mbox{stop}$}\\r_j+\gamma\max_{a'}\widehat{Q}(s_{j+1},a';\phi)& \mbox{ otherwise.}\end{array}\right.$
%   \IF{$a_j=\mbox{stop}$}
%   \STATE Set $y_j=r_j$
%   \ELSE
%   \STATE Set $y_j=r_j+\gamma\max_{a'}\widehat{Q}(s_{j+1},a';\phi)$
%   \ENDIF
   \STATE Perform gradient descent on $\left(y_j-Q(s_j,a_j;\theta)\right)^2$ with respect to network parameters $\theta$
   \STATE Every $U$ episodes reset target network $\widehat{Q}=Q$
   \ENDFOR
   \label{alg:ddql}
\end{algorithmic}
\end{algorithm}

%\EDcomments{
\newpage
\subsection{Customized Categorical distributional RL Algorithm}\label{app:algc51}

\begin{algorithm}[h]%[tb]
   \caption{Customized Categorical distributional RL Algorithm, a.k.a. C51 when $N=51$}
   \label{alg:doubleC51Learning}
\begin{algorithmic}[1]
   \STATE {\bfseries Inputs:} A set of $M$ episodes $\{\V{s}_{0:T}^i\}_{i=1}^M$, $V_{max}$ and $V_{min}$ are the maximum and minimum values of possible returns, $N$ is the number of atom probabilities 
   \STATE {\bfseries Initialize:} replay memory $D$ to capacity $C$ of episodes
  \STATE {\bfseries Initialize:} discrete support $z_{k}=V_{min} + k \Delta z$ for $k=0$ to $N-1$, with $\Delta z=\frac{V_{max} - V_{min}}{N-1}$
   \STATE {\bfseries Initialize:} value distribution $P$ with random weights $\theta$, and target distribution $\widehat{P}$'s with weights $\phi=\theta$ 
   \FOR{episode $i =1$ {\bfseries to} $M$}
   \STATE Perform lines \ref{alg:searchA}-\ref{alg:searchC} from Algorithm \ref{alg:doubleQLearning} where line \ref{alg:searchB} uses $Q(s_t^i,a)=\sum_{k} z_{k}P_{k}(s_t^i,a;\theta)$
   \STATE Sample a random mini-batch of buffers $B$ of sequence transitions $(s_j, a_j, r_j, s_{j+1})$ from $D$
   %%%%%%%%%%%
%   \STATE \textcolor{gray}{\#$N=51$ is the number of atom probabilities $\{z_{i}=V_{min} + i \Delta z\}$ where $\Delta z=\frac{V_{max} - V_{min}}{N-1}$}
%   \STATE \textcolor{gray}{\#$V_{max}$ and $V_{min}$ are the maximum and minimum values of possible returns, respectively}
   \STATE Set $m_{k}=0$, for all $k \in 1,\dotsc,N-1$
   %\STATE for $j \in 1,\dotsc,N-1$ do
   \FOR{$k=1$ {\bfseries to} $N-1$}
%   \STATE \textcolor{gray}{\#Compute the projection of $\hat{\mathcal{T}}z_{k}$ onto the support $\{z_{i}\}$}
%\State \#Project distributional Bellman update onto the support $\{z_{k}\}_{k=0}^{N-1}$
\STATE $\#$ Project distributional Bellman update onto the support $\{z_{k}\}_{k=0}^{N-1}$
   \STATE Set $v_k= \max(V_{min},\min(V_{max},r_{j} + \gamma z_{k}))$
   \STATE $\#$ Identify support index of projection
   \STATE Set $b_{k} = (v_{k} - V_{min}) / \Delta z$ 
   \STATE Set $l = \lfloor b_{k}\rfloor, u = \lceil b_{k}\rceil, a'=\arg\max_a \sum_{k} z_{k}P_{k}(s_{j+1},a;\phi)$
%  \STATE \textcolor{gray}{\#Distribute probability of sample}
\STATE $\#$ Distribute probability of sample
   \STATE Set $m_{l} = m_{l} + \widehat{P}_{k}(s_{j+1},a';\phi)(u-b_{k})$, set $m_{u} = m_{u} + \widehat{P}_{k}(s_{j+1},a';\phi)(b_{k}-l)$ 
   \ENDFOR
   \STATE Perform a gradient step on cross-entropy loss $-\sum_{k}m_{k}\log P_{k}(s_{j},a_{j};\theta)$ with respect to $\theta$
   %%%%%%%%%%%
   \STATE Every $U$ episodes update target distribution $\widehat{P}=P$, i.e.  $\phi=\theta$%=P(\cdot,\cdot,\theta)$
   \ENDFOR
   \label{alg:c51}
\end{algorithmic}
\end{algorithm}
%}

\section{Final choice of hyper-parameters}\label{app:hyper}

\begin{table}[h]
\centering
\caption{Hyperparameters of the different RL versions.}
\label{tab:hyperparams-table}
\begin{tabular}{@{}ccl@{}}
\toprule
Task                     & Algorithm & \multicolumn{1}{c}{Hyperparameters}                                          \\ \midrule
\multirow{3}{*}{GBM}     & DDQN      & learning-rate=0.0001  + batch-size=128 + $C$=10000  + $U$=300  \\ \cmidrule(l){2-3} 
                         & C51       & learning-rate=0.0025  + batch-size=64  + $C$=3000  +  $U$=30  \\ \cmidrule(l){2-3} 
                         & IQN       & learning-rate=0.00005 + batch-size=128 + $C$=3000  + $U$=1000 \\ \bottomrule
\multirow{3}{*}{S\&P500} & DDQN      & learning-rate=0.005  +  batch-size=64  + $C$=10000  + $U$=300 \\ \cmidrule(l){2-3} 
                         & C51       & learning-rate=0.0025  +  batch-size=64  + $C$=3000  +  $U$=30 \\ \cmidrule(l){2-3} 
                         & IQN       & learning-rate=0.0025  + batch-size=64  + $C$=3000  + $U$=100  \\ \midrule
\end{tabular}
\end{table}

\removed{
\newpage
\section{Categorical distributional RL Algorithm}\label{app:algc51}

\begin{algorithm}[h]%[tb]
   \caption{Double Categorical distributional RL Algorithm}
   \label{alg:doubleC51Learning}
\begin{algorithmic}[1]
   \STATE {\bfseries Inputs:} A set of $M$ episodes $\{\V{s}_{0:T}^i\}_{i=1}^M$
   \STATE {\bfseries Initialize:} replay memory $D$ to capacity $C$ of episodes
   \STATE {\bfseries Initialize:} discrete value distribution function $P$ with random weights $\theta$
   \STATE {\bfseries Initialize:} target discrete value distribution function $\widehat{P}$ with weights $\phi=\theta$
   \FOR{episode $i =1$ {\bfseries to} $M$}
   \STATE {\bfseries Initialize:} episode buffer $B$ to capacity $T$
   \STATE With probability $\epsilon$ episode is random and a uniformly random day to stop $t_{stop}$ is selected
   \FOR{$t=0$ {\bfseries to} $T$}
   \IF{episode is random}
   %\STATE Set $y_j=r_j$
   %%%%%%%%
   \STATE Set $a_t=\left\{\begin{array}{cl}\mbox{continue}&\mbox{ if $t<t_{stop}$}\\\mbox{stop}&\mbox{ otherwise.}\end{array}\right.$
   %%%%%%%%
   \ELSE
   \STATE Set $a_t=\arg\max_a{Q(s_t^i,a)}$ where $Q(s_t^i,a)=\sum_{j} z_{j}P_{j}(s_t^i,a;\theta)$
   \ENDIF
   %%%
   %\STATE select $a_t=\arg\max_a{Q(s_t,a;\theta)}$
   \STATE Execute action $a_t$ and observe reward $r_t$, store transition $(s_t^i, a_t, r_t, s_{t+1}^i)$ in $B$
   \IF{($a_t=\mbox{stop}$ or $t=t_{stop}$)}% and $K-x_{k+t-1}>0$}
   \STATE Exit \textbf{for} loop
   \ENDIF
   \ENDFOR
   \STATE Store episode buffer $B$ in replay memory $D$. If $D$ full, drop the oldest episode
   \STATE Sample a random mini-batch of buffers $B$ of sequence transitions $(s_j, a_j, r_j, s_{j+1})$ from $D$
   %%%%%%%%%%%
   \STATE \textcolor{gray}{\#$N=51$ is the number of atom probabilities $\{z_{i}=V_{min} + i \Delta z\}$ where $\Delta z=\frac{V_{max} - V_{min}}{N-1}$}
   \STATE \textcolor{gray}{\#$V_{max}$ and $V_{min}$ are the maximum and minimum values of possible returns, respectively}
   \STATE $m_{k}=0$, $k \in 1,\dotsc,N-1$
   %\STATE for $j \in 1,\dotsc,N-1$ do
   \FOR{$k=1$ {\bfseries to} $N-1$}
   \STATE \textcolor{gray}{\#Compute the projection of $\hat{\mathcal{T}}z_{k}$ onto the support $\{z_{i}\}$}
   \STATE $\hat{\mathcal{T}}z_{k} \leftarrow [r_{j} + \gamma z_{k}]_{V_{min}}^{V_{max}}$
   \STATE $b_{k} \leftarrow (\hat{\mathcal{T}}z_{k} - V_{min}) / \Delta z$ \textcolor{gray}{\#$b_{k} \in [0,N-1]$}
   \STATE $l \leftarrow \floor{b_{k}}, u \leftarrow \ceil{b_{k}}, a'=\arg\max_a{\widehat{Q}(s_{j+1},a)}$
   \STATE \textcolor{gray}{\#Distribute probability of $\hat{\mathcal{T}}z_{k}$}
   \STATE $m_{l} \leftarrow m_{l} + \widehat{P}_{k}(s_{j+1},a';\phi)(u-b_{k})$
   \STATE $m_{u} \leftarrow m_{u} + \widehat{P}_{k}(s_{j+1},a';\phi)(b_{k}-l)$
   \ENDFOR
   \STATE Perform gradient descent on cross-entropy loss $-\sum_{i}m_{i}\log P_{i}(s_{j},a_{j};\theta)$ with respect to network parameters $\theta$
   %%%%%%%%%%%
   \STATE Every $U$ episodes reset target network $\widehat{P}=P$
   \ENDFOR
   \label{alg:c51}
\end{algorithmic}
\end{algorithm}

}

\end{document}